\documentclass[10pt,twocolumn,letterpaper]{article}

\usepackage{cvpr}
\usepackage{times}
\usepackage{epsfig}
\usepackage{graphicx}
\usepackage{amsmath}
\usepackage{amssymb}
\usepackage[font={footnotesize},skip=5pt]{caption}
\usepackage[font={footnotesize},justification=centering]{subcaption}
\usepackage{physics}
\usepackage{array}
\usepackage{multirow}

\usepackage{cite}

\DeclareMathOperator*{\argmax}{arg\,max}

\graphicspath{ {images/} }

\usepackage[breaklinks=true,bookmarks=false]{hyperref}

\cvprfinalcopy 

\def\cvprPaperID{2318} 

\begin{document}

\pagenumbering{gobble}

\title{Compare and Contrast: Learning Prominent Visual Differences}

\author{Steven Chen \qquad Kristen Grauman\\
The University of Texas at Austin\\
}

\maketitle

\begin{abstract}
Relative attribute models can compare images in terms of all detected properties or attributes, exhaustively predicting which image is fancier, more natural, and so on without any regard to ordering. However, when humans compare images, certain differences will naturally stick out and come to mind first. These most noticeable differences, or \emph{prominent differences}, are likely to be described first. In addition, many differences, although present, may not be mentioned at all. In this work, we introduce and model prominent differences, a rich new functionality for comparing images. We collect instance-level annotations of most noticeable differences, and build a model trained on relative attribute features that predicts prominent differences for unseen pairs. We test our model on the challenging UT-Zap50K shoes and LFW10 faces datasets, and outperform an array of baseline methods. We then demonstrate how our prominence model improves two vision tasks, image search and description generation, enabling more natural communication between people and vision systems.
\end{abstract}

\section{Introduction} \label{introduction}

\begin{figure}
    \centering
    \captionsetup[subfigure]{justification=centering,font=footnotesize,labelfont=footnotesize}
    \includegraphics[width=0.8\linewidth]{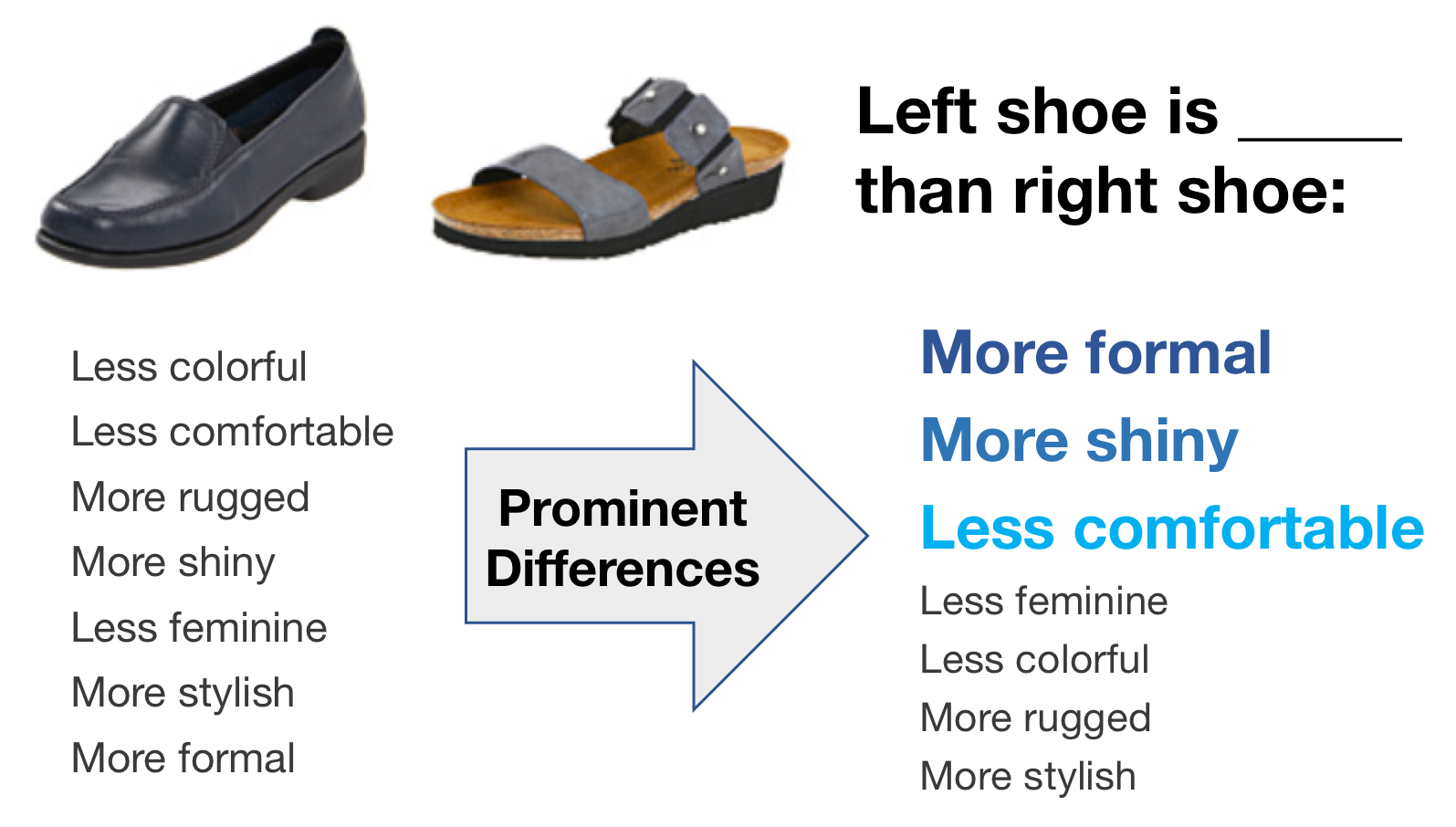}
    \caption[Prominent Differences]{When we compare images, certain differences stick out over others. Although all the attribute statements on the left are true, the prominent differences in bold stand out and would often be described first, whereas the others may not even be mentioned.}
    \label{fig:prominentdifferencesdefinition}
\end{figure}

Suppose you are asked to compare and contrast two different shoes, shown in Figure \ref{fig:prominentdifferencesdefinition}. You might say that the left shoe is \textit{more formal} than the right shoe, then perhaps state that the left shoe is \textit{more shiny} and \textit{less comfortable} than the right shoe. As soon as you are given the images, these differences stick out and are most noticeable. However, consider that the two shoes have a huge number of differences. For instance, the left shoe is \textit{more rugged} than the right shoe, and also \textit{darker}. Although these other differences are certainly present and true, they are much less noticeable to us, and we would likely mention them later, or not at all.

In general, when we perform any comparison task on a pair of images, certain differences stick out as being most noticeable out of the space of all discernible differences. These most noticeable differences, or \textit{prominent differences}, stand out and would be described first, while most other differences are not as noticeable and would typically not be mentioned in a description.

In this work, we introduce and learn prominent differences in images, expressing them through relative attributes. When people compare images, they can describe differences in their \textit{attributes}, human-nameable visual properties of images~\cite{describingobjects, zeroshot, whittlesearch, simile, relativeattributes, dominance, deepfashion} used to describe anything from materials (\textit{smooth}, \textit{furry}) and parts (\textit{has leg}, \textit{has glasses}) to styles (\textit{sporty}, \textit{formal}) and expressions (\textit{smiling}, \textit{sad}). \textit{Relative attributes}, or attributes that indicate an image's attribute strength relative to other images, provide an intuitive and meaningful representation for visual comparison, and have been widely used for vision tasks~\cite{virality, whittlesearch, partattribute, finegrained, jnd, deeprelative, deeprelative2, deeprelative3, spokenattributes}. Relative attributes express comparisons of attribute strength (\eg, image X is \textit{smiling more} than Y, but \textit{smiling less} than Z), and are the natural vocabulary of our proposed prominent differences.

Prominent differences have many practical applications in vision. Humans interact with vision systems as both users and supervisors, and naturally communicate prominence. For instance, in an interactive search task, where humans provide comparative feedback (\eg, I would like to see images like this shoe, but \textit{more formal}~\cite{whittlesearch}), the attributes that people elect to comment on are prominent differences. In zero-shot learning with relative attributes~\cite{zeroshot,relativeattributes}, where humans describe unseen visual categories to a machine by comparing with seen categories, prominence could enhance learning by better understanding these comparisons. Prominent differences are the properties humans provide first when making comparisons, and thus directly influence how humans interpret comparison descriptions. Prominence could also be used to highlight the differences that stick out to people between fashion styles, extending recent work that uses attributes for fashion description~\cite{forecastingfashion, deepfashion,unsupervisedfashion}. 

\begin{figure}
    \centering
    \captionsetup[subfigure]{justification=centering,font=footnotesize,labelfont=footnotesize}
    
    \begin{subfigure}[c]{0.3\linewidth}
        \centering
        \includegraphics[width=0.475\linewidth]{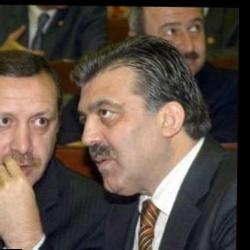}
        \includegraphics[width=0.475\linewidth]{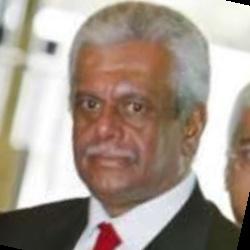}
        \caption{dark hair (\textgreater)}
        \label{fig:example1}
    \end{subfigure}
    \hfill
    \begin{subfigure}[c]{0.3\linewidth}
        \centering
        \includegraphics[width=0.475\linewidth]{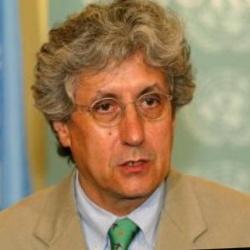}
        \includegraphics[width=0.475\linewidth]{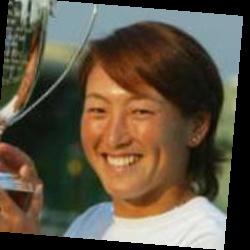}
        \caption{smiling (\textless)}
        \label{fig:example2}
    \end{subfigure}
    \hfill
    \begin{subfigure}[c]{0.3\linewidth}
        \centering
        \includegraphics[width=0.475\linewidth]{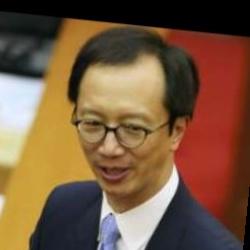}
        \includegraphics[width=0.475\linewidth]{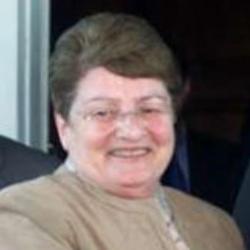}
        \caption{forehead (\textgreater)}
        \label{fig:example3}
    \end{subfigure}
    
    \vspace{0.25cm}
    
    \begin{subfigure}[c]{0.3\linewidth}
        \centering
        \includegraphics[width=0.475\linewidth]{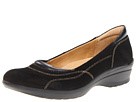}
        \includegraphics[width=0.475\linewidth]{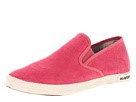}
        \caption{colorful (\textless)}
        \label{fig:example4}
    \end{subfigure}
    \hfill
    \begin{subfigure}[c]{0.3\linewidth}
        \centering
        \includegraphics[width=0.475\linewidth]{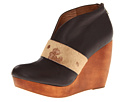}
        \includegraphics[width=0.475\linewidth]{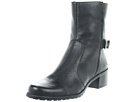}
        \caption{fancy (\textgreater)}
        \label{fig:example5}
    \end{subfigure}
    \hfill
    \begin{subfigure}[c]{0.3\linewidth}
        \centering
        \includegraphics[width=0.475\linewidth]{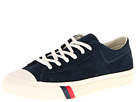}
        \includegraphics[width=0.475\linewidth]{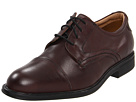}
        \caption{formal (\textless)}
        \label{fig:example6}
    \end{subfigure}
    
    \vspace{0.1cm}
    
    \caption[Prominent Differences in Image Pairs]{Different attributes stand out as prominent for different image pairs. According to the consensus of seven MTurk judges, even though \ref{fig:example1} and \ref{fig:example2} both differ in darkness of hair, \textit{dark hair} sticks out as most prominent in \ref{fig:example1} but not in \ref{fig:example2}. The wide color difference makes \textit{colorful} prominent in \ref{fig:example4}, while in \ref{fig:example6}, a combination of properties result in \textit{formal} as the prominent difference.}
    \label{fig:prominentdifferencesintro}
\end{figure}

Modeling prominent differences is challenging due to several key reasons. First, there is a large variety of reasons why an attribute stands out as prominent for any image pair. For instance, large differences in attribute strength can play a role (Figure \ref{fig:example4}), as well as absence of other significant differences (Figure \ref{fig:example1}) and unusual occurrences (Figure \ref{fig:example3}). In general, complex interactions between the attributes of two images cause certain differences to stand out. Second, humans use a large and diverse vocabulary when expressing prominence, which a model should support. Finally, prominent differences are observed between individual images. As we will show, simply predicting prominence based on the prior frequency of usage for the attribute words is not sufficient. Thus, prominent differences must be modeled at the image instance level.

In this work, we propose to model prominent differences. We collect a novel dataset of prominent difference annotations, propose a model based on relative attribute features learned with deep spatial transformer networks or large margin rankers, and evaluate on two unique and challenging domains: the UT-Zap50K shoes dataset~\cite{finegrained} and the LFW10 faces dataset~\cite{relativeparts}. We show that our model significantly outperforms an array of baselines for predicting prominent differences, including an adaptation of the state-of-the-art binary attribute dominance approach~\cite{dominance}. Finally, we demonstrate how our prominence model can be used to enhance two vision applications: interactive image search and description generation.

\section{Related Work}

\noindent \textbf{Attributes} \hspace{1em}  Attributes are semantic and machine-understandable properties (\eg, \textit{smiling}, \textit{shiny}) that are used by people to describe images~\cite{importance, attributediscovery, virality, describingobjects, zeroshot, attributeshades, whittlesearch, simile, deeplearningface, partattribute, relativeattributes, sunattribute, spokenattributes, relativeparts, imagerankingmulti, dominance, finegrained, jnd, interestingness, babytalk, referit, humansintheloop, deepfashion, clothing, forecastingfashion, unsupervisedfashion}. Attributes serve as expressive mid-level features for recognition~\cite{sunattribute, simile, deeplearningface}. Attributes have also been used as a vocabulary for learning unseen visual categories, known as zero-shot learning~\cite{zeroshot, betweenclassattribute, dominance, relativeattributes}. Attributes are well suited for applications in fashion, where they have been used to forecast style popularity \cite{forecastingfashion}, organize outfits by style \cite{unsupervisedfashion}, and drive interactive search~\cite{fashionsearch}. Recently, deep convolutional neural networks (CNNs) have shown improved attribute prediction accuracy over previous approaches in various fields such as faces~\cite{deeplearningface} and fashion~\cite{deepfashion}. In contrast to previous work, which focuses on detecting the presence of attributes in images, we learn which particular attribute differences stick out when comparing images.

\vspace{0.25cm}

\noindent \textbf{Relative Attributes} \hspace{1em} Relative attributes, first introduced in~\cite{relativeattributes}, represent an image's attribute strength with respect to other images~\cite{virality, robustsubjective, whittlesearch, partattribute, finegrained, jnd, deeprelative, deeprelative2, deeprelative3, spokenattributes, semanticjitter}, and are a richer representation than binary presence/absence. Relative attributes enable \textit{visual comparisons} between images (\eg, the left shoe is \textit{more sporty} than the right), and have been used to discern fine-grained differences~\cite{finegrained, jnd} and predict image virality~\cite{virality}. Recently, deep CNNs have been used to both predict relative attributes~\cite{deeprelative, deeprelative2, deeprelative3} as well as generate synthetic images of varying attribute strengths~\cite{semanticjitter, attributetoimage}. However, no prior work considers which relative attributes stand out, or what relative attributes humans tend to use in speech. Our work introduces prominent differences, a novel functionality representing most noticeable differences in the vocabulary of relative attributes.

\vspace{0.25cm}

\noindent \textbf{Importance of Objects and Attributes} \hspace{1em} Different concepts of visual importance have used attributes~\cite{importance, objectimportance, dominance, interestingness}.  Attributes have been used to learn object importance (defined in \cite{objectimportance} as the likelihood an object is named first), which can be used to help rank image aesthetics~\cite{interestingness, deepaesthetics}. As opposed to objects, we consider the separate concept of prominent differences, which are selected from a vocabulary of linguistic properties.

Turakhia and Parikh~\cite{dominance} introduce the related concept of binary attribute dominance, measuring which binary attributes (\eg, \textit{is 4 legged}, \textit{not smiling}) stand out more for different object categories. Our work is distinct: we learn which \textit{relative differences} stick out \textit{between two images}, while \cite{dominance} learns which \textit{binary attributes} are more apparent \textit{by category}. For instance, given the category of sneakers, \cite{dominance} may detect general trends of \textit{sporty} and \textit{comfortable}; however, our prominence approach captures important differences between image instances, such as one specific shoe being prominently \textit{more rugged} than another. We later extend binary attribute dominance and compare with our approach (\textit{cf}. Section \ref{baselines}).

\vspace{0.25cm}

\noindent \textbf{Image Saliency} \hspace{1em} Works modeling saliency (\eg,~\cite{wherehumanslook, deepsaliency, oldsaliency}) have used attributes to predict what regions people tend to look at in images. Although saliency may have an influence on prominence, it refers to low-level regions in single images, whereas prominence is a linguistic, pairwise concept, and the result of a combination of mid-level cues.

\vspace{0.25cm}

\noindent \textbf{Image Search} \hspace{1em} Image search has benefited from attribute-based approaches~\cite{imagerankingmulti, whittlesearch, dominance, jnd}.  Attribute queries have been used in search~\cite{imagerankingmulti}, and improvements have been made using binary attribute ordering~\cite{dominance}. We use relative attribute ordering, and apply to the interactive WhittleSearch~\cite{whittlesearch, whittlesearch2}. Whereas previous work solicits richer user feedback for WhittleSearch~\cite{jnd}, we leverage the implicit prominence choices users make to obtain better results with no additional human input or interface changes.

\vspace{0.25cm}

\noindent \textbf{Describing Images} \hspace{1em} As semantic properties, attributes are well-suited for visual description~\cite{describingobjects, relativeattributes, spokenattributes, dominance, babytalk, referring, referit}. Recent work uses attributes to generate binary attribute descriptions~\cite{describingobjects, babytalk, dominance}. Attributes have been used to generate referring expressions~\cite{referring, referit}, phrases identifying specific objects in an image. In comparison, our work focuses on attribute differences, which could be used to improve referring expressions. Works have generated image descriptions using relative attribute comparisons~\cite{relativeattributes, spokenattributes}. However, these methods list exhaustive statements in arbitrary order; we improve these by focusing on differences that are prominent and natural to state.

\section{Approach} \label{approach}

First, we present an overview of relative attribute models (Section \ref{relativeattributemodels}). Next, we introduce our approach for modeling prominent differences (Section \ref{modeling}). Finally, we detail how we gather our prominence datasets (Section \ref{annotating}).

\subsection{Background: Relative Attribute Models} \label{relativeattributemodels}

\begin{figure}
    \centering
    
    \includegraphics[width=0.115\linewidth]{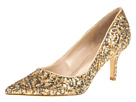}
    \includegraphics[width=0.115\linewidth]{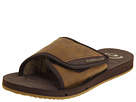}
    \includegraphics[width=0.115\linewidth]{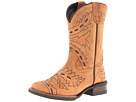}
    \includegraphics[width=0.115\linewidth]{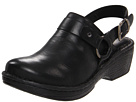}
    \includegraphics[width=0.115\linewidth]{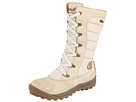}
    \includegraphics[width=0.115\linewidth]{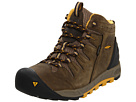}
    \includegraphics[width=0.115\linewidth]{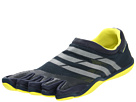}
    \includegraphics[width=0.115\linewidth]{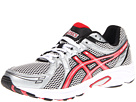}
    
    \includegraphics[width=\linewidth]{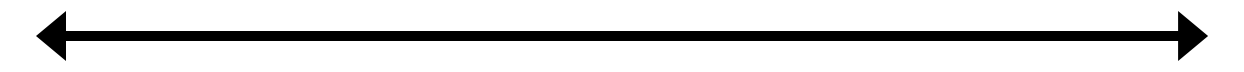}
    
    \vspace{-0.1cm}
    
    less sporty \hfill more sporty
    
    \vspace{0.25cm}
    
    \includegraphics[width=0.115\linewidth]{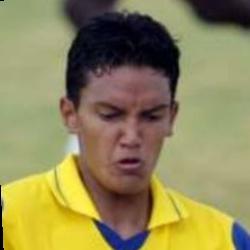}
    \includegraphics[width=0.115\linewidth]{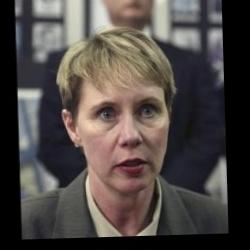}
    \includegraphics[width=0.115\linewidth]{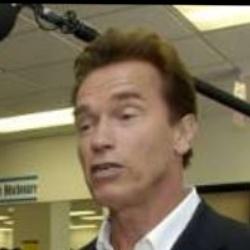}
    \includegraphics[width=0.115\linewidth]{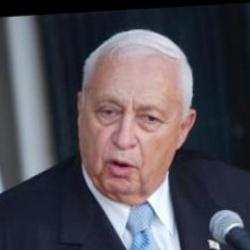}
    \includegraphics[width=0.115\linewidth]{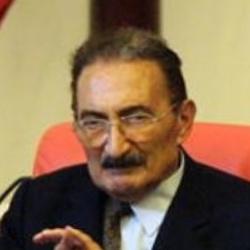}
    \includegraphics[width=0.115\linewidth]{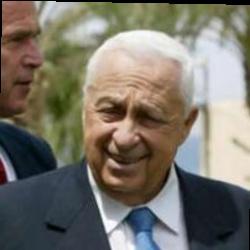}
    \includegraphics[width=0.115\linewidth]{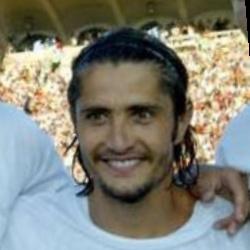}
    \includegraphics[width=0.115\linewidth]{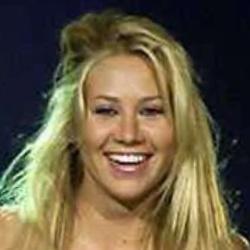}
    
    \includegraphics[width=\linewidth]{Arrow}
    
    \vspace{-0.1cm}
    
    less smiling \hfill more smiling
    
    \vspace{0.1cm}
    
    \caption[Relative Attributes]{Relative attributes can rank images across a range of strengths.}
    \label{fig:rel}
\end{figure}

Relative attributes are semantic visual properties that represent the strength of an attribute in an image relative to other images~\cite{relativeattributes}. While binary attributes represent just the presence or absence of a property (\eg, \textit{is smiling}, \textit{is not smiling}), relative attributes rank images based on their attribute strength scores (see Figure \ref{fig:rel}), and are thus well suited for visual comparison.

We now describe a general framework for relative attribute rankers. Suppose we have a set of images $I = \{i\}$, along with a vocabulary of $M$ relative attributes $A = \{a_m\}_{m=1}^M$. Let $x_i \in \mathbb{R}^D$ represent the image's \textit{D}-dimensional descriptor, which could be comprised of GIST~\cite{gist}, color, part descriptors, CNN features, or just raw pixels. Given an image pair $y_{ij} = (x_i, x_j)$, the goal of the ranker is to determine if one image contains more of $a_m$ than the other, or if both images have similar strengths of $a_m$.

Relative attribute models use sets of labeled image pairs for supervised learning~\cite{relativeattributes, robustsubjective, partattribute, deeprelative, deeprelative2, deeprelative3}. The model is given a set of ordered pairs $O_m = \{(i, j)\}$ and a set of unordered pairs $S_m = \{(i, j)\}$ such that $(i, j) \in O_m \Longrightarrow i > j$, \ie, $i$ contains more of $a_m$ that $j$, and $(i, j) \in S_m \Longrightarrow i \sim j$, \ie, $i$ and $j$ have similar strengths of $a_m$.

Relative attribute models learn a ranking function $\mathcal{R}_m(x_i)$ for each attribute $a_m$ to best satisfy the constraints:
\begin{align}
    \forall (i, j) \in O_m : \mathcal{R}_m(x_i) > \mathcal{R}_m(x_j) \\
    \forall (i, j) \in S_m : \mathcal{R}_m(x_i) = \mathcal{R}_m(x_j).
\end{align}

Different learning objectives are used to quantify constraint satisfaction, such as a wide margin classification objective~\cite{relativeattributes} for an SVM~\cite{ranksvm} ranker, or a RankNet objective~\cite{deeprelative, deeprelative2, deeprelative3, ranknet} for a deep CNN ranker. We experiment with both such models in our implementation.

\vspace{0.22cm}

\noindent \textbf{Ranking SVM} \hspace{1em} Ranking SVM rankers optimize $\mathcal{R}_m^{(svm)}(x_i) = w_m^T x_i$ to preserve ordering while maximizing distance between closest points when projected onto $w_m$. $w_m \in \mathbb{R}^D$ is the weight vector to be learned, and is linear here; nonlinear models are also possible using kernels. Ranking SVMs have seen wide use for relative attributes \cite{relativeattributes, virality, robustsubjective, whittlesearch, partattribute, finegrained}.

\vspace{0.22cm}

\noindent \textbf{CNN Ranker} \hspace{1em} Deep CNN based rankers have emerged as a strong alternative for predicting relative attributes~\cite{deeprelative, deeprelative2, deeprelative3}. These models generally use a CNN optimized for paired ranking loss~\cite{ranknet}. Compared to the Ranking SVM models, they typically achieve better accuracy but require more time for training and larger quantities of training data. We use Singh and Lee's Siamese network~\cite{deeprelative} in our work, where each branch consists of a spatial transformer network and ranker network.

\subsection{Modeling Prominent Differences} \label{modeling}

We now introduce our approach for modeling prominent differences, defined as the attribute a person would mention first when comparing two given images. We first introduce a naive method, then present our approach.

\vspace{0.22cm}

\noindent \textbf{Naive Widest Relative Difference} \hspace{1em} Given an image pair $y_{uv}$, a simple approach for predicting prominence would be to directly compare their relative attribute scores $r_m^u = \mathcal{R}_m(x_u)$ and $r_m^v = \mathcal{R}_m(x_v)$. After normalizing these scores, one can compute the relative difference $\abs{r_m^u - r_m^v}$, for how different the images are in terms of $a_m$. By taking the maximum
\begin{align}
    \mathcal{W}^{uv} = \argmax_m \abs{r_m^u - r_m^v}
\end{align}
over all attributes, we obtain the naive \textit{widest relative difference} $\mathcal{W}^{uv}$.

We hypothesize $\mathcal{W}^{uv}$ alone is inadequate for predicting prominence. As illustrated in Section \ref{introduction}, there are several contributing factors to prominence, such as unusual occurrence of attributes, and many other complex interactions between the properties of the image pair. In the next section, we describe our approach for modeling prominence, which uses a novel representation to learn the interactions that cause prominence. Later, we demonstrate the efficacy of our approach through comparison with widest differences and other methods (\textit{cf}. Section \ref{baselines}).

\vspace{0.22cm}

\noindent \textbf{Our Approach} \hspace{1em} Suppose we have a set of images $I = \{x_i\}$, along with $M$ relative attributes $A = \{a_m\}_{m=1}^M$ as defined before. In addition, for each $a_m$, we are given a set of unordered prominence image pairs $U_m = \{(x_i, x_j)\}$ such that the most prominent difference between $x_i$ and $x_j$ is $a_m$. Note that $U_m$ is distinct from $O_m$ and $S_m$, the relative attribute pairs used to train the relative attribute ranker.

Our goal is to construct a model that, given a novel pair of images $y_{uv} = (x_u, x_v)$, predicts which attribute $\mathcal{A}^{uv}$ is the most prominent difference for that image pair.

First, in order to represent $y_{ij}$ as an unordered pair, we need a symmetric transformation $\phi(y_{ij}) = \phi(y_{ji})$ that combines the attributes of the images into a joint representation, such that the model always predicts the same prominent difference for each specific pair. The representation should also capture the wide variety of factors for why certain properties stand out as prominent, so that the model may effectively learn.

To create our feature representation $\phi$, we first compute the relative attribute scores $r_m^i = \mathcal{R}_m(x_i)$ for both images $x_i$ in the pair and all attributes in the vocabulary, using the models described in Section \ref{relativeattributemodels}, resulting in scores $r_1^i, \dotsc, r_M^i$ for each image. We then compute $\phi$ as the average of the pair's scores for each attribute, and concatenate the absolute difference between the pair's attribute scores, creating a feature vector of length $2M$:

{\footnotesize
\begin{align}
    \phi(y_{ij}) = (\frac{r_1^i + r_1^j}{2}, \dotsc, \frac{r_M^i + r_M^j}{2}, \abs{r_1^i - r_1^j}, \dotsc, \abs{r_M^i - r_M^j}).
\end{align}
}%

This feature representation captures the individual relative attribute properties while maintaining symmetry: for instance, unordered pair scores for each attribute $a_m$ can be reconstructed from two feature vector components $\phi(y_{ij})_m \pm \frac{1}{2} \phi(y_{ij})_{M+m}$. We standardize attribute scores to zero mean and unit variance before they are input into $\phi(y_{ij})$.

We experimented with other $\phi$ transformations on the attribute scores of an image pair, including element-wise products, absolute difference, and weighted averages. We select the given formulation of $\phi$ due to its strong performance in practice.

Given this representation, we now build $M$ predictors $\mathcal{P}_m(y_{uv})$ for $m = 1, \dotsc, M$ such that $\mathcal{P}_m(y_{uv})$ is the predicted confidence score that the prominent difference for $y_{uv}$ is $a_m$. We predict the prominence confidence for each attribute $a_m$ using
\begin{align}
    \mathcal{P}_m(y_{uv}) = \mathcal{S}_m (w_m^T \phi(y_{uv}))
\end{align}
where $w_m^T$ are the weights learned of a binary linear classifier, and $\mathcal{S}_m$ is a function mapping classifier outputs to confidence scores.

\begin{figure}[t]
    \centering
    \includegraphics[width=0.9\linewidth]{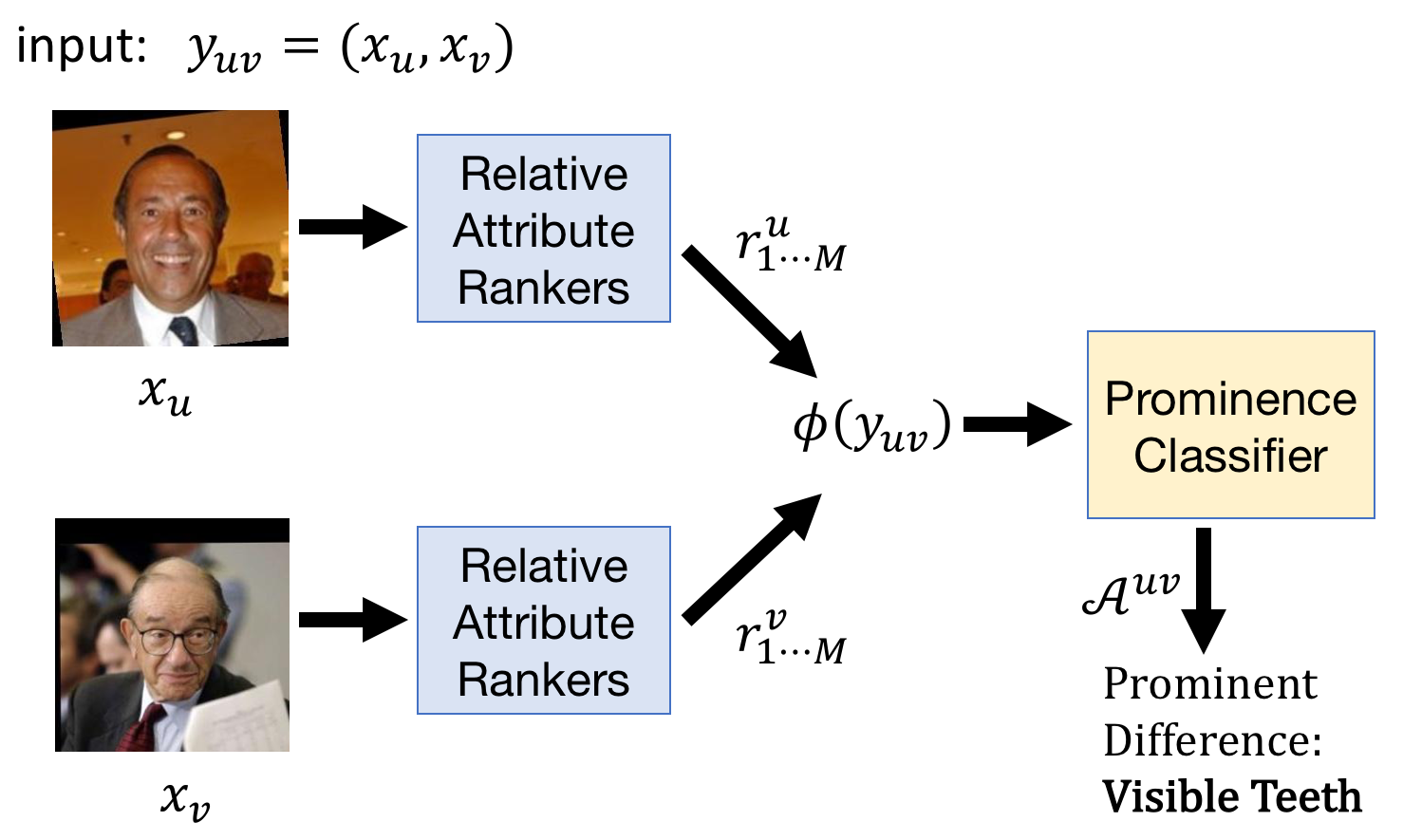}
    \caption[Pipeline for Prominent Difference Prediction]{Our approach pipeline for prominent difference prediction.}
    \label{fig:pipeline}
\end{figure}

To learn the classifier weights $w_m^T$ for each $a_m$, we mark all training pairs from $U_m$ as positive examples, and all other training pairs as negatives. We use a linear SVM classifier for each attribute, due to its strong performance in practice. We use Platt's method~\cite{plattscaling} ($\mathcal{S}_m$) to transform each classifier output into posterior probabilities.

Using our full predictors $\mathcal{P}_{1 \dotsc M}$, we predict the most prominent difference $\mathcal{A}^{uv}$ for $y_{uv}$ by choosing the attribute with the highest confidence:
\begin{align}
    \mathcal{A}^{uv} = \argmax_m(\mathcal{P}_m(y_{uv}))).
\end{align}

In addition, we can also return the top $k$ prominent differences by selecting the $k$ attributes with the highest confidence scores. This can be used to generate a description for a pair of images describing their $k$ most prominent differences (\textit{cf}. Section \ref{descriptiongeneration}).

Our model follows the structure of a one vs.~all multiclass classifier, with our relative attribute features $\phi(y_{uv})$ as input features and the prominent difference as class labels. Other models could certainly be considered, such as a one vs.~one classifier, or a ranker. We choose the one vs.~all classifier over a one vs.~one for several reasons: its strong prediction performance, its easy interpretability for individual attributes, and its efficiency (only requiring one classifier per attribute). We choose a classifier approach vs.~a ranker for its ease of collecting natural human perception of prominence (see the next section), as opposed to exhaustive comparisons between all combinations of attributes, which is less intuitive and can lead to noisier results. Deep models could certainly also fit into our framework, although they require substantially more manual annotation.

\subsection{Annotating Prominent Differences} \label{annotating}

No dataset exists for training or evaluating prominent differences, so we collect human annotations of prominence for image pairs at the instance level using Mechanical Turk.

To collect human perception of prominent differences, we first create a large and diverse vocabulary of $M$ attributes that generally stuck out to annotators when viewing the dataset (see Section \ref{datasets} for vocabulary details). We then show an annotator a pair of randomly selected images, along with a list of all $M$ attributes $\{a_m\}, m \in \{1, \dotsc, M\}$, and ask which attribute out of the list sticks out as the most noticeable difference for that image pair.

It is important to highlight that we ask each annotator to select \textit{just one} prominent difference. This allows annotators to provide their natural first impression. Additionally, we provide the entire vocabulary of $M$ attributes to choose from, which aids in ensuring that at least a subset of choices are noticeably different for almost all image pairs. 

In addition, our approach is scalable: it requires only one annotation question per image pair, regardless of the number of attributes in the vocabulary, vs. $\binom{M}{2}$ combinations of attribute pair questions required to annotate one instance of binary dominance in~\cite{dominance}. This helps us scale to the instance level and capture fine-grained information on which specific images and features lead to prominence, whereas~\cite{dominance} collects data at the category-level, projecting the same values to all instance images in a category and losing valuable instance-level characteristics in the process.

To obtain our prominent difference ground truth labels, we collect annotations from seven annotators for each image pair. Then, for each pair, we rank the attributes by frequency chosen, and label the highest ranked attribute as the ground truth most prominent difference.

\section{Applications of Prominent Differences}

We now present our approaches for applying prominent differences to two human-centric applications, image search (Section \ref{imagesearch}) and description generation (Section \ref{descriptiongeneration}).

\subsection{Image Search} \label{imagesearch}

First, we consider applying prominent differences to WhittleSearch~\cite{whittlesearch, whittlesearch2}, an interactive image search framework where users provide relative attribute feedback through comparisons (\eg, I would like images that are \textit{more formal} than reference image X).

WhittleSearch intersects the relative attribute constraints $c_1, \dotsc, c_n$ provided by the user, ranking database images by how many constraints they satisfy. In each search iteration, the user is shown a page of top ranked images and selects reference images and relative attribute constraints on those images. WhittleSearch then adds the new feedback to the set of all constraints and ranks images accordingly.

\begin{figure}
    \centering
    \includegraphics[width=0.9\linewidth]{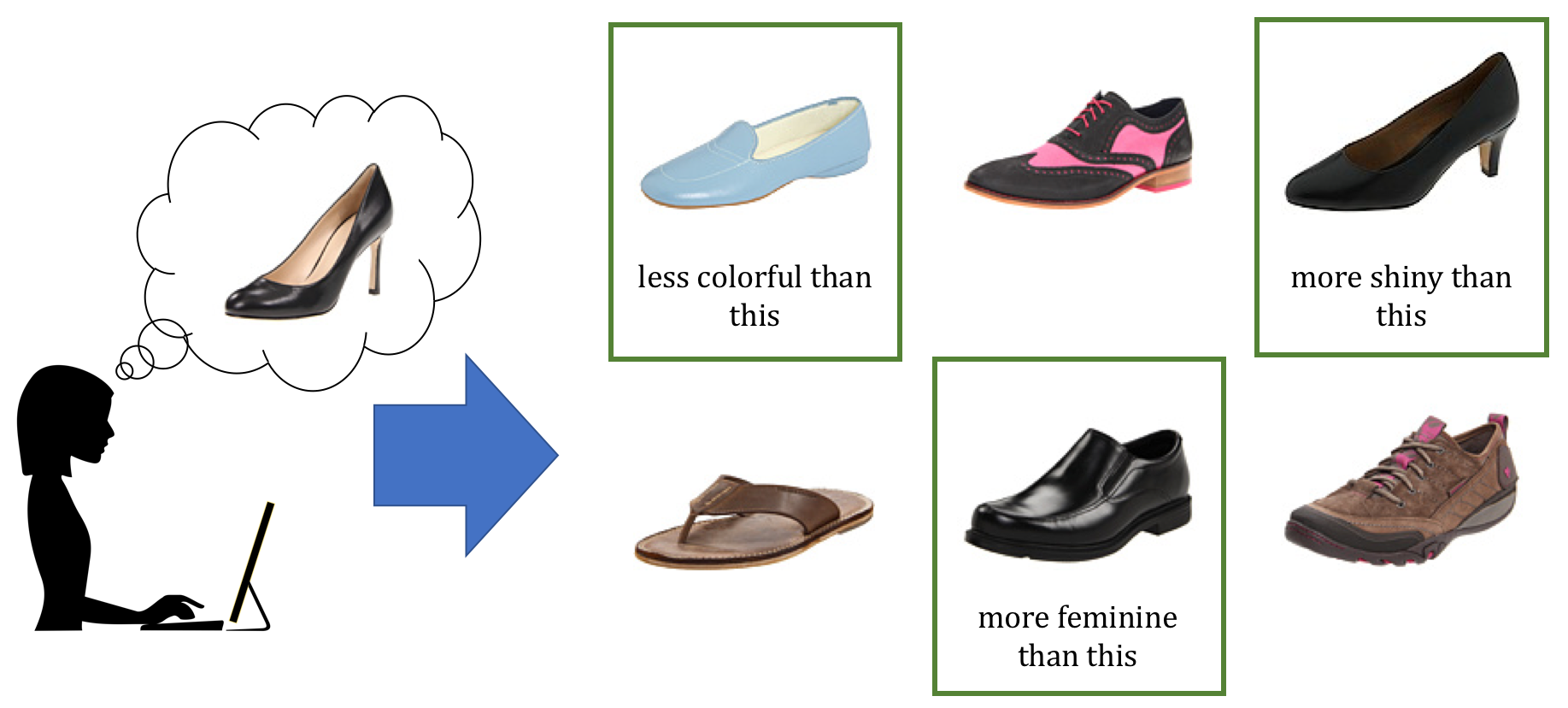}
    
    \caption[WhittleSearch Relative Attribute Feedback]{In WhittleSearch~\cite{whittlesearch, whittlesearch2}, a user chooses reference images and constraints from the search page. We hypothesize that users will provide \textit{prominent differences} between the reference and their target as feedback, and show how the proposed method can improve search.}
    \label{fig:whittlesearchmotivation}
\end{figure}

When users provide feedback in the form of ``What I am looking for is \textit{more/less} $a_m$ than image $x_{ref}$", by definition, they will provide prominent differences between the reference image $x_{ref}$ and their mental target (see Figure \ref{fig:whittlesearchmotivation}). Thus, images are more likely to be relevant if they are \textit{prominently} different in $a_m$ with $x_{ref}$. We model this by introducing a relevance term $p$ for each database image $x_i$ using prominence:
\begin{align}
    p(x_i \mid c_1, \dotsc, c_n) \propto \prod_{c} \mathcal{P}_{m_c}(x_i, x_{ref_c}),
\end{align}
where $\mathcal{P}_{m_c}$ is our prominence predictor for $m_c$, and attribute $m_c$ and reference $ref_c$ are the constraint parameters from constraint $c$, for all $c_1, \dotsc, c_n$.

Then, we rank images \textit{within each group satisfying the same number of constraints} by listing them in descending order of $p(x_i)$. We use this approach to maintain the overall constraint ordering of WhittleSearch, while using prominence to significantly improve the ordering of images that share the same number of satisfied constraints. 

A strength of our approach is that it does not require any additional user input: we simply use the user's existing feedback for prominence data. As we will show, our approach is especially impactful in the first several iterations of search, when many images satisfy all or most of the feedback constraints, and would be otherwise randomly ordered by the existing WhittleSearch algorithm~\cite{whittlesearch}.

\subsection{Description Generation} \label{descriptiongeneration}

For our second application, we consider applying prominent differences to generate textual image descriptions. Given a novel image pair, we want to generate a description comparing the images in terms of their attributes (\eg, Image X is \textit{more sporty} and \textit{less formal} than Image Y).

When humans compare images, by definition, they state prominent differences first. In addition, humans will not name all differences for a pair; instead, they will usually describe a subset of the prominent differences. Previous work~\cite{relativeattributes, spokenattributes} generates descriptions comparing images in terms of all attributes, in an arbitrary order. We argue that this is not sufficient; listing out all differences is too lengthy, while listing a random subset can miss key differences.

We propose generating descriptions containing prominent differences. Namely, given a novel image pair $y_{uv}$, we sort all attributes in descending order of their prominence scores, and generate a description with the top $k$ prominent differences. For example, given two shoe images, our model can generate the description ``The left shoe is \textit{more sporty}, \textit{less stylish}, and \textit{less shiny} than the right shoe," stating the three most prominent differences between the instance images.

\section{Results} \label{results}

We first introduce the two datasets we use (Section \ref{datasets}), followed by the baselines that we compare our approach to (Section \ref{baselines}). Finally, we evaluate our approach on prominence prediction (Section \ref{evaluation}), as well as on image search (Section \ref{imagesearchresults}) and description generation (Section \ref{descriptionresults}).

\subsection{Datasets} \label{datasets}

We now introduce the two datasets used in our experiments, then highlight annotator agreement statistics.

\vspace{0.25cm}

\noindent \textbf{UT-Zap50K Shoes Dataset} \hspace{1em} The UT-Zap50K Dataset~\cite{finegrained} is a dataset of 50,025 shoe images from Zappos. We use a vocabulary of ten relative attributes for our experiments: (1) \textit{sporty}, (2) \textit{comfortable}, (3) \textit{shiny}, (4) \textit{rugged}, (5) \textit{fancy}, (6) \textit{colorful}, (7) \textit{feminine}, (8) \textit{tall}, (9) \textit{formal}, (10) \textit{stylish}. These were selected from data collected in \cite{semanticjitter}, in which annotators were asked to provide the first difference that comes to mind. We use the 19 provided categories (\eg, AnkleBoots, OxfordsShoes, \etc) as categories for the binary dominance baseline~\cite{dominance}.

We randomly sample 2,000 images, and collect prominence for 4,990 sample pairs. For the SVM ranker and binary dominance, we generate CNN features from the fc7 layer of AlexNet~\cite{alexnet}. For the deep CNN ranker~\cite{deeprelative}, the inputs are raw pixels. For prominence prediction, we report the average of 10-fold cross validation. Images used in training pairs are disjoint from images used in testing pairs.

\vspace{0.22cm}

\noindent \textbf{LFW10 Faces Dataset} \hspace{1em} The LFW10 Dataset~\cite{relativeparts} is a collection of 2,000 face images from LFW~\cite{lfw}, along with 10,000 relative attribute annotations over ten different attributes, (1) \textit{bald head}, (2) \textit{dark hair}, (3) \textit{eyes open}, (4) \textit{good looking}, (5) \textit{masculine}, (6) \textit{mouth open}, (7) \textit{smiling}, (8) \textit{visible teeth}, (9) \textit{visible forehead}, (10) \textit{young}. We use these attributes as our vocabulary, and create categories for binary dominance~\cite{dominance} by matching images to their subjects. We keep all individuals with three or more instances, resulting in 1,064 images from 150 categories.

We collect prominence for 1,463 sample pairs. For image descriptors for the SVM rankers and binary dominance, we use the 8,300 dimension part features learned on dense SIFT~\cite{sift} provided by \cite{relativeparts}. We reduce the dimensionality to 200 using PCA to avoid overfitting. We report the average of 5-fold cross validation.

\vspace{0.22cm}

\noindent \textbf{Annotator Agreement} \hspace{1em} 77\% of Zap50K image pairs had three or more annotators out of seven agree on the most prominent difference, with 87\% for LFW10. On average, 3.8 unique attributes were chosen as most noticeable for each image pair for Zap50K, with 3.3 for LFW10. (See Supp for additional attribute frequency statistics.) This high level of agreement shows that prominent differences are in fact consistent for most comparisons.

\subsection{Baselines} \label{baselines}

\noindent \textbf{Binary Attribute Dominance \cite{dominance}} \hspace{1em} Our first baseline is Turakhia and Parikh's binary attribute dominance~\cite{dominance}. We follow the authors' approach as closely as possible, collecting separate dominance and binary attribute annotations for training the model (see Supp for details). We convert relative attributes into binary equivalents (\eg, \textit{sportiness} becomes \textit{is sporty} and \textit{is not sporty}), and extend their model to handle prominence by first computing binary dominance for each attribute and image in a pair, then selecting the attribute with the highest score for an individual image as the predicted prominent difference.

\vspace{0.22cm}

\noindent \textbf{Widest Relative Difference \texttt{+} tf-idf} \hspace{1em} This baseline uses the (standardized) widest relative difference $\mathcal{W}^{uv}$ described in Section \ref{modeling} to estimate prominence. We add tf-idf weighting to this baseline, which we find is a strict improvement over no weighting scheme.

\vspace{0.22cm}

\noindent \textbf{Single Image Prominence} \hspace{1em} This baseline is trained on the same prominence labels as our approach, but projects the labels onto the two images of each pair. It then trains a multiclass SVM with individual images as inputs, as opposed to our approach, which trains on pairs and a pairwise feature representation.

\vspace{0.22cm}

\noindent \textbf{Prior Frequency} \hspace{1em} The final baseline we use is a simple ``prior frequency" model, which predicts prominent differences proportionally according to their frequency of occurrence in the ground truth.

\subsection{Prominent Difference Evaluation} \label{evaluation}

\begin{figure}[t]
    \centering
    \begin{subfigure}[c]{0.49\linewidth}
        \centering
        \includegraphics[width=\linewidth]{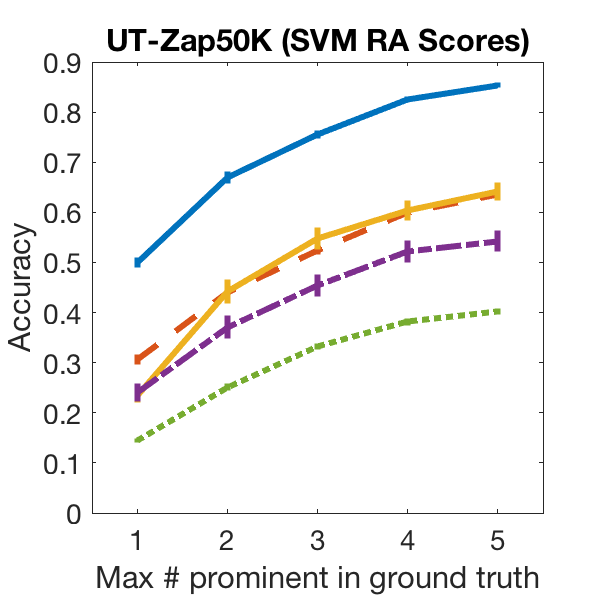}
        \label{fig:zapsvm}
    \end{subfigure}
    \hfill
    \begin{subfigure}[c]{0.49\linewidth}
        \centering
        \includegraphics[width=\linewidth]{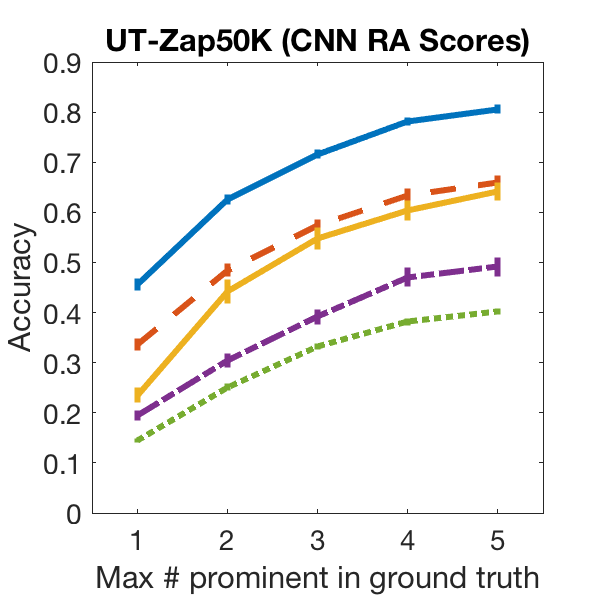}
        \label{fig:zapcnn}
    \end{subfigure}
    
    \vspace{-0.2cm}
    
    \includegraphics[width=0.95\linewidth]{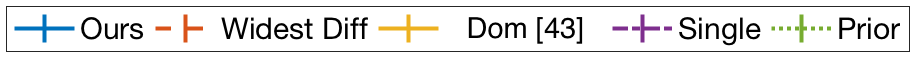}
    
    \vspace{0.1cm}
    
    \begin{subfigure}[c]{0.49\linewidth}
        \centering
        \includegraphics[width=\linewidth]{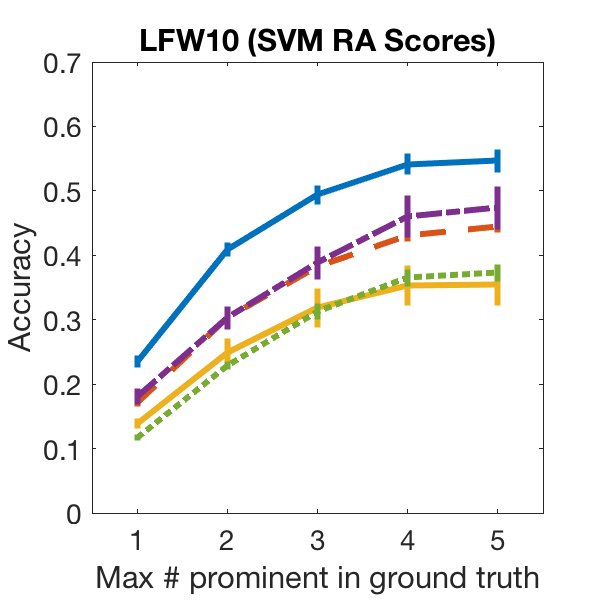}
        \label{fig:lfwsvm}
    \end{subfigure}
    \hfill
    \begin{subfigure}[c]{0.49\linewidth}
        \centering
        \includegraphics[width=\linewidth]{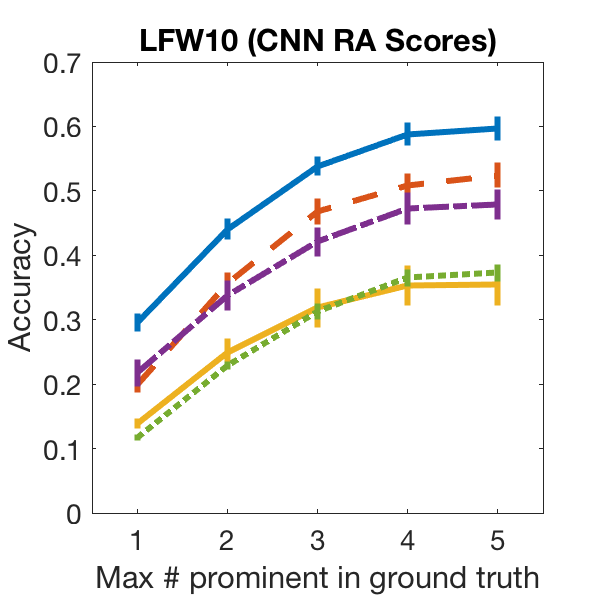}
        \label{fig:lfwcnn}
    \end{subfigure}
    
    \vspace{-0.2cm}
    
    \caption[Prominence Evaluation Accuracy]{Prominence prediction accuracy results for two datasets (top, bottom) and either SVM or CNN-based attribute rankers (left, right).}
    \label{fig:eval}
\end{figure}

We evaluate prominent difference prediction accuracy, with each model predicting a single most prominent difference for each image pair. Recall that seven annotators supply ground truth prominence on each image pair. Because there is not always a unanimous prominent difference, we evaluate accuracy over a range of $k$ maximum attributes marked as ground truth correct, to account for variance in perception. Specifically, we sort attributes by their frequency chosen, creating a partial ranking of $c$ attributes, and take the $\text{min}(k, c)$ most chosen as the prominence ground truth. We mark a pair as correct if the prediction $\mathcal{A}^{uv}$ is present in the ground truth. At $k = 1$, only the most prominent attribute is considered correct.

Figure \ref{fig:eval} shows accuracy results. We divide results for both Zap50K and LFW10 into two plots, one for each attribute ranker used (ranking SVM and deep CNN). Our approach significantly outperforms all baselines for prediction. We observe sizable gains of roughly 20-22\% on Zap50K, and 6-15\% on LFW10 over the strongest baselines. This clearly demonstrates the advantage of our approach, which uses pairwise relative attribute features to learn the interactions between visual properties that result in prominent differences.

\begin{figure}[t]
    \centering
    \captionsetup[subfigure]{justification=centering,font=footnotesize,labelfont=footnotesize}
    
    \begin{subfigure}[c]{0.3\linewidth}
        \centering
        \includegraphics[width=0.475\linewidth]{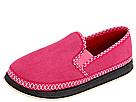}
        \includegraphics[width=0.475\linewidth]{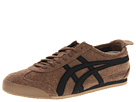}
        \caption{\textbf{color (\textgreater)},\\ sporty, comfort}
        \label{fig:zap1}
    \end{subfigure}
    \hfill
    \begin{subfigure}[c]{0.3\linewidth}
        \centering
        \includegraphics[width=0.475\linewidth]{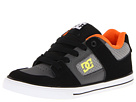}
        \includegraphics[width=0.475\linewidth]{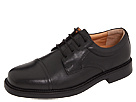}
        \caption{\textbf{sporty (\textgreater)},\\ color, comfort}
        \label{fig:zap2}
    \end{subfigure}
    \hfill
    \begin{subfigure}[c]{0.3\linewidth}
        \centering
        \includegraphics[width=0.475\linewidth]{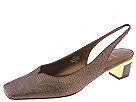}
        \includegraphics[width=0.475\linewidth]{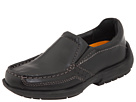}
        \caption{\textbf{feminine (\textgreater)},\\ comfort, shiny}
        \label{fig:zap6}
    \end{subfigure}
    
    \vspace{0.25cm}
    
    \begin{subfigure}[c]{0.3\linewidth}
        \centering
        \includegraphics[width=0.475\linewidth]{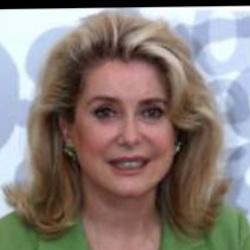}
        \includegraphics[width=0.475\linewidth]{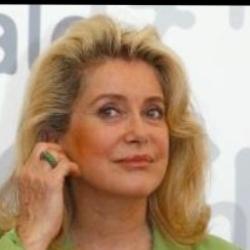}
        \caption{\textbf{teeth (\textgreater)},\\ mouth open, smiling}
        \label{fig:lfw6}
    \end{subfigure}
    \hfill
    \begin{subfigure}[c]{0.3\linewidth}
        \centering
        \includegraphics[width=0.475\linewidth]{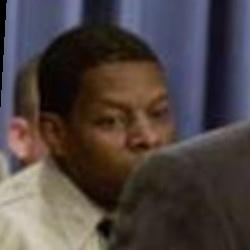}
        \includegraphics[width=0.475\linewidth]{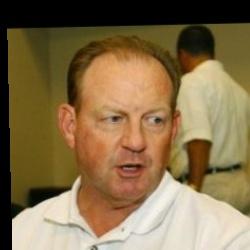}
        \caption{\textbf{bald head (\textless)},\\ dark hair, teeth}
        \label{fig:lfw2}
    \end{subfigure}
    \hfill
    \begin{subfigure}[c]{0.3\linewidth}
        \centering
        \includegraphics[width=0.475\linewidth]{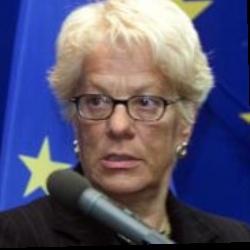}
        \includegraphics[width=0.475\linewidth]{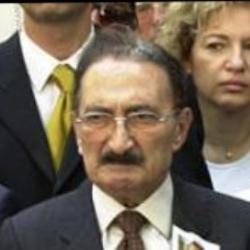}
        \caption{\textbf{dark hair (\textless)},\\ mouth open, smiling}
        \label{fig:lfw3}
    \end{subfigure}
    
    \vspace{0.1cm}
    
    \caption[Sample Prominent Difference Prediction, Strong Results]{Prominence predictions made by our approach. Predicted most prominent attribute in bold, followed by next two most confident attributes.}
    \label{fig:good}
\end{figure}

Our results show that the baselines are not adequate for predicting prominent differences. For widest relative difference, its lower accuracy demonstrates that strength difference is only one contributing factor to prominence: our approach is able to effectively capture other important causes. We also outperform binary dominance~\cite{dominance} significantly. The weak performance of the single image prominence baseline demonstrates that prominence is a pairwise phenomenon, requiring both images for context.

Comparing the use of the two attribute rankers, both yield similar performance on Zap50K but we benefit from the CNN ranker scores on LFW10. The advantages of our approach hold whether using SVM rankers or deep rankers. It is important to highlight that our contributions are orthogonal to the choice of attribute ranker: our approach can learn from relative attribute scores generated from any model.

Figure \ref{fig:good} shows qualitative examples. For example, the shoes in \ref{fig:zap1} are very different in many properties; despite this, our model accurately predicts \textit{colorful} as the most prominent difference. Although the images in \ref{fig:lfw6} are of the same person, our model is able to accurately predict \textit{visible teeth} as the most prominent difference. (See Supp for more examples.)

\subsection{Image Search Results} \label{imagesearchresults}

\begin{figure}[t]
    \centering
    \begin{subfigure}[c]{0.49\linewidth}
        \centering
        \includegraphics[width=\linewidth]{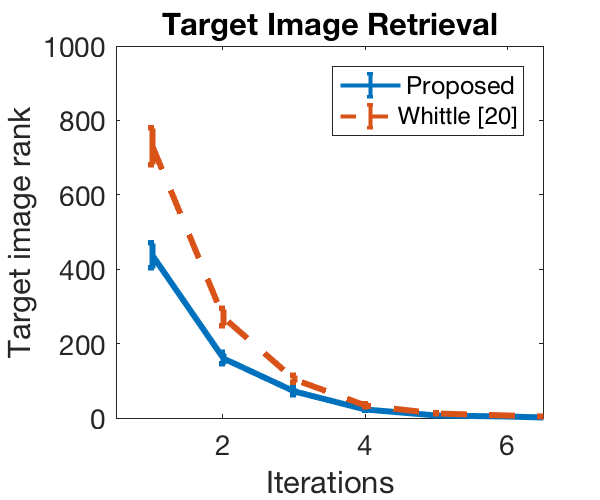}
        \label{fig:whittlegraph}
    \end{subfigure}
    \hfill
    \begin{subfigure}[c]{0.49\linewidth}
        \centering
        \includegraphics[width=\linewidth]{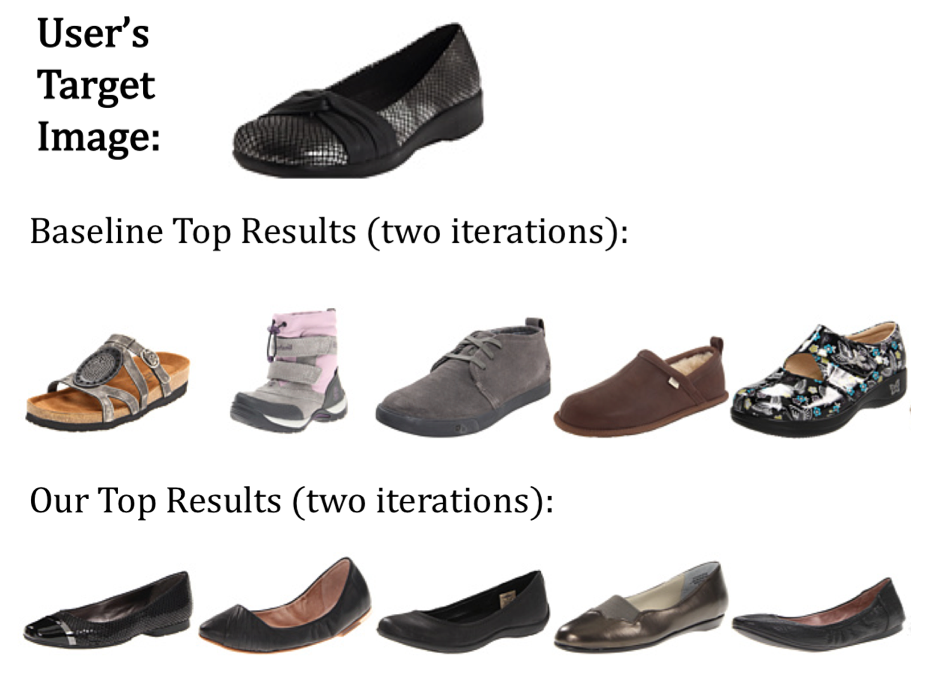}
        \label{fig:whittleexample}
    \end{subfigure}
    
    \vspace{-0.3cm}
    
    \caption[Image Search Results]{Image search results. Quantitative results on left, lower rank is better. On right, qualitative search results with a user's target image, followed by baseline results and our results. Our approach returns significantly more similar images to the user's mental target.}
    \label{fig:whittlesearchresults}
\end{figure}

For our image search application on WhittleSearch~\cite{whittlesearch, whittlesearch2}, we evaluate a proof-of-concept experiment using the Zap50K dataset. We use Zap50K for its large size, and sample 5,000 unseen images as our experimental database.

Due to the cost of obtaining human feedback for each combination of image and results list, we generate feedback automatically following~\cite{whittlesearch}. A random subset of images from the top results page is chosen as reference images. For the user's feedback between the target $x_t$ and each reference image $x_{ref}$, the user selects the most prominent difference $\mathcal{A}^{t,ref}$ to provide feedback upon. To simulate variance in human perception, we add noise by randomly selecting 75\% of feedback as prominent differences, and 25\% as random true differences. We select 200 images as the user's mental targets. At each iteration, the user is shown the top 16 results, selects 8 images as references, and provides 8 feedback constraints using the references.

Figure \ref{fig:whittlesearchresults} shows the results (more in Supp). Our approach substantially improves the target image rank over the first several iterations of search, and returns significantly more relevant images, without requiring any additional feedback.

\subsection{Description Generation Results} \label{descriptionresults}

\begin{figure}[t]
    \centering
    \begin{subfigure}[c]{0.46\linewidth}
        \centering
        \includegraphics[width=\linewidth]{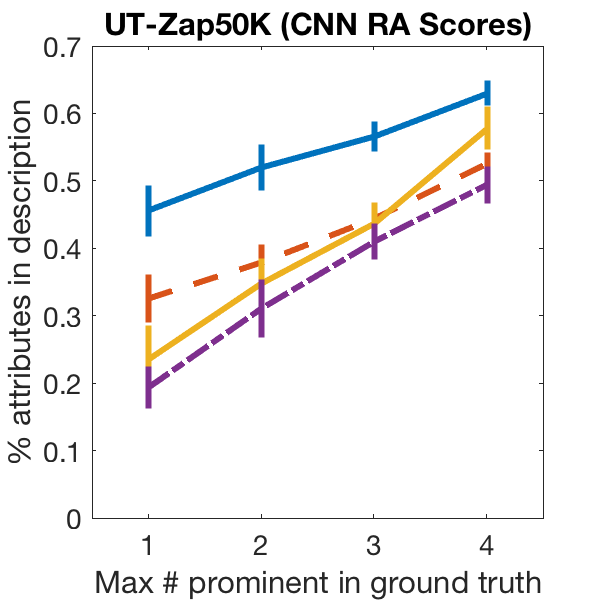}
        \label{fig:desczapsvm}
    \end{subfigure}
    \hfill
    \begin{subfigure}[c]{0.46\linewidth}
        \centering
        \includegraphics[width=\linewidth]{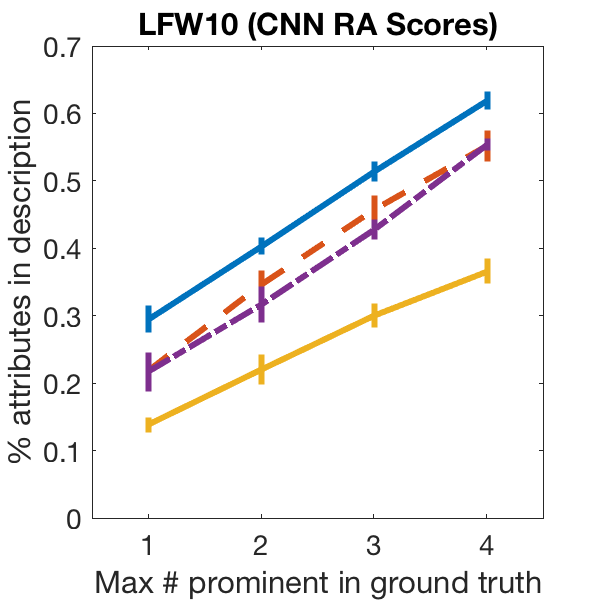}
        \label{fig:desczapcnn}
    \end{subfigure}
    
    \vspace{-0.2cm}
    
    \includegraphics[width=0.75\linewidth]{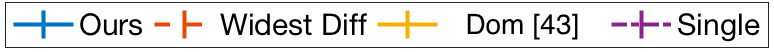}
    
    \caption[Description Generation Accuracy]{Description generation offline results.}
    \label{fig:desc}
\end{figure}

We evaluate generated descriptions in one offline and one online experiment. For the offline experiment, we output the top $k$ most prominent attributes that would be in a generated description, and check what percentage of the $k$ ground truth prominent attributes are present, a metric that is critical to description quality. We compare our approach to the three strongest baselines and report results with the CNN attribute ranker in Figure \ref{fig:desc} (see Supp for similar results with other ranker). Our approach outperforms all baselines, generating descriptions capturing more human-perceived prominent differences. 

For the online experiment, we ask annotators to judge generated descriptions. Specifically, we present an image pair and two descriptions---our description of predicted prominent differences, and a baseline description with randomly chosen \textit{true} differences---and ask which description is more natural and appropriate. We sample 200 pairs from Zap50K and 100 pairs from LFW10, generate descriptions with three statements each, and have seven judges provide feedback per pair, taking the majority vote.

For Zap50K, 69\% of people preferred our description, compared to 31\% for the baseline, with a p-value $< 0.0001$,  with 61\% and 39\% for LFW10, with a p-value of $0.01$, respectively. We also ran the same experiment using annotator ground truth prominence vs. the same baseline: people preferred the ground truth description 69\% of the time for Zap50K and 70\% for LFW10 (see Table \ref{table:desc}). Our generated Zap50K descriptions are closer to the ground truth performance than for LFW10 due to our method's higher prominence prediction accuracy on Zap50K. These results demonstrate that describing images using prominent differences results in significantly more natural descriptions.

\begin{figure}
    \centering
    \captionsetup[subfigure]{justification=centering,font=footnotesize,labelfont=footnotesize}
    
    \begin{subfigure}[c]{0.475\linewidth}
        \centering
        \includegraphics[width=0.35\linewidth]{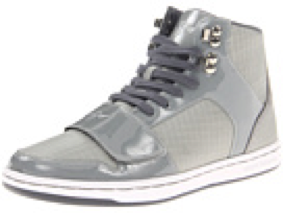}
        \includegraphics[width=0.35\linewidth]{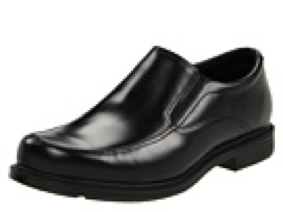}
        \caption{Left is less shiny, less formal, and more colorful than the right.}
        \label{fig:good1}
    \end{subfigure}
    \hfill
    \begin{subfigure}[c]{0.475\linewidth}
        \centering
        \includegraphics[width=0.35\linewidth]{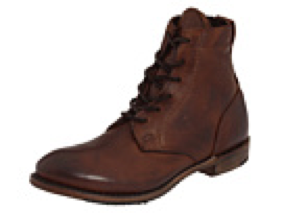}
        \includegraphics[width=0.35\linewidth]{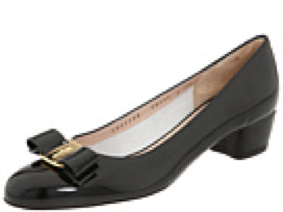}
        \caption{Left is less feminine, more rugged, and less shiny than the right.}
        \label{fig:good2}
    \end{subfigure}
    
    \vspace{0.15cm}
    
    \begin{subfigure}[c]{0.475\linewidth}
        \centering
        \includegraphics[width=0.35\linewidth]{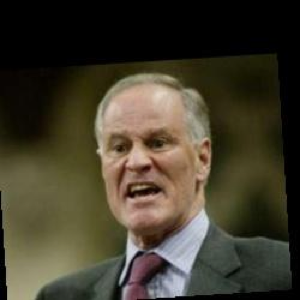}
        \includegraphics[width=0.35\linewidth]{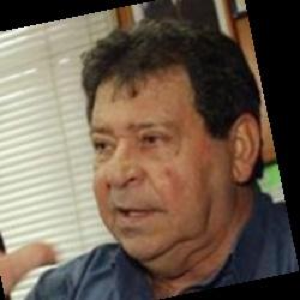}
        \caption{Left has less dark hair, more bald head, and more mouth open than the right.}
        \label{fig:good5}
    \end{subfigure}
    \hfill
    \begin{subfigure}[c]{0.475\linewidth}
        \centering
        \includegraphics[width=0.27\linewidth]{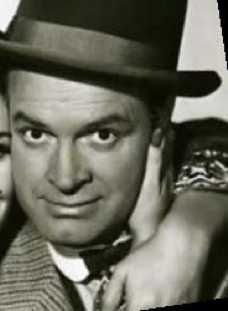}
        \includegraphics[width=0.35\linewidth]{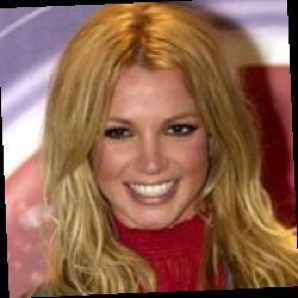}
        \caption{Left is more masculine, less smiling, and less visible teeth than the right.}
        \label{fig:good6}
    \end{subfigure}
    
    \vspace{0.1cm}
    
    \caption[Sample Textual Descriptions]{Sample generated descriptions by proposed approach.}
    \label{fig:descexamples}
\end{figure}

\begin{table}
\centering
\resizebox{0.9\linewidth}{!}{%
\begin{tabular}{|c||l  c|l  c|} \hline
\multirow{2}{5em}{Zap50K} & Ours: & 69\% & Baseline: & 31\% \\ \cline{2-5}
& Ground Truth: & 69\% & Baseline: & 31\% \\ \hline
\multirow{2}{5em}{LFW10} & Ours: & 61\% & Baseline: & 39\% \\ \cline{2-5}
& Ground Truth: & 70\% & Baseline: & 30\% \\ \hline
\end{tabular}}
\caption[Description Generation Study Results]{Description generation human study results.}
\label{table:desc}
\end{table}

\section{Conclusion}

We introduced prominent differences, a novel functionality for comparing images. When humans compare images, certain prominent differences naturally stick out, while others, although present, may not be mentioned. We present a novel approach for modeling prominence at the image pair level. Experimental results on the UT-Zap50K and LFW10 datasets show that our proposed approach significantly outperforms an array of baseline methods for predicting prominent differences. In addition, we demonstrate how prominence can be used to improve two applications: interactive image search and description generation.

There is strong potential for future work. In zero-shot learning using relative attributes~\cite{relativeattributes}, where a human supervisor teaches a machine about an unseen visual category using comparisons, humans will provide prominent differences as supervision. This knowledge, if integrated, could result in improved classification without requiring any additional human effort. In addition, prominent differences could be used to improve referring expressions~\cite{referring, unambiguousdescription}, phrases identifying specific objects in an image. Prominent differences could be used to identify best differences to help distinguish one object over others.

\vspace{0.2cm}

\noindent \textbf{Acknowledgements} \hspace{1em} We thank Aron Yu for helpful discussions. This research is supported in part by NSF IIS-1065390 and an Amazon Research Award.

\bibliographystyle{ieee}
\bibliography{egbib}

\begin{thebibliography}{10}\itemsep=-1pt

\bibitem{forecastingfashion}
Z.~Al-Halah, R.~Stiefelhagen, and K.~Grauman.
\newblock {F}ashion {F}orward: {F}orecasting {V}isual {S}tyle in {F}ashion.
\newblock In {\em ICCV}, 2017.

\bibitem{importance}
A.~Berg, T.~Berg, H.~Daume, J.~Dodge, A.~Goyal, X.~Han, A.~Mensch, M.~Mitchell,
  A.~Sood, K.~Stratos, and K.~Yamaguchi.
\newblock {U}nderstanding and {P}redicting {I}mportance in {I}mages.
\newblock In {\em CVPR}, 2012.

\bibitem{attributediscovery}
T.~Berg, A.~Berg, and J.~Shih.
\newblock {A}utomatic {A}ttribute {D}iscovery and {C}haracterization from
  {N}oisy {W}eb {D}ata.
\newblock In {\em ECCV}, 2010.

\bibitem{humansintheloop}
S.~Branson, C.~Wah, F.~Schroff, B.~Babenko, P.~Welinder, P.~Perona, and
  S.~Belongie.
\newblock {V}isual {R}ecognition with {H}umans in the {L}oop.
\newblock In {\em ECCV}, 2010.

\bibitem{ranknet}
C.~Burges, T.~Shaked, E.~Renshaw, A.~Lazier, M.~Deeds, N.~Hamilton, and
  G.~Hullender.
\newblock {L}earning to {R}ank using {G}radient {D}escent.
\newblock In {\em ICML}, 2005.

\bibitem{clothing}
H.~Chen, A.~Gallagher, and B.~Girod.
\newblock {D}escribing {C}lothing by {S}emantic {A}ttributes.
\newblock In {\em ECCV}, 2012.

\bibitem{virality}
A.~Deza and D.~Parikh.
\newblock Understanding image virality.
\newblock In {\em CVPR}, 2015.

\bibitem{interestingness}
S.~Dhar, V.~Ordonez, and T.~L. Berg.
\newblock {H}igh {L}evel {D}escribable {A}ttributes for {P}redicting
  {A}esthetics and {I}nterestingness.
\newblock In {\em CVPR}, 2011.

\bibitem{describingobjects}
A.~Farhadi, I.~Endres, D.~Hoiem, and D.~Forsyth.
\newblock {D}escribing {O}bjects by their {A}ttributes.
\newblock In {\em CVPR}, 2009.

\bibitem{robustsubjective}
Y.~Fu, T.~M. Hospedales, T.~Xiang, J.~Xiong, S.~Gong, Y.~Wang, and Y.~Yao.
\newblock Robust subjective visual property prediction from crowdsourced
  pairwise labels.
\newblock {\em CoRR}, abs/1501.06202, 2015.

\bibitem{unsupervisedfashion}
W.~Hsiao and K.~Grauman.
\newblock {L}earning the {L}atent {L}ook: {U}nsupervised {D}iscovery of a
  {S}tyle-{C}oherent {E}mbedding from {F}ashion {I}mages.
\newblock In {\em ICCV}, 2017.

\bibitem{lfw}
G.~B. Huang, M.~Ramesh, T.~Berg, and E.~Learned-Miller.
\newblock Labeled faces in the wild: A database for studying face recognition
  in unconstrained environments.
\newblock Technical Report 07-49, University of Massachusetts, Amherst, 2007.

\bibitem{oldsaliency}
L.~Itti, C.~Koch, and E.~Niebur.
\newblock A model of saliency-based visual attention for rapid scene analysis.
\newblock {\em PAMI}, 20(11):1254--1259, 1998.

\bibitem{zeroshot}
D.~Jayaraman and K.~Grauman.
\newblock Zero-shot recognition with unreliable attributes.
\newblock In {\em ECCV}, 2014.

\bibitem{ranksvm}
T.~Joachims.
\newblock {O}ptimizing {S}earch {E}ngines {U}sing {C}lickthrough {D}ata.
\newblock In {\em SIGKDD}, 2002.

\bibitem{wherehumanslook}
T.~Judd, K.~Ehinger, F.~Durand, and A.~Torralba.
\newblock {L}earning to {P}redict {W}here {H}umans {L}ook.
\newblock In {\em ICCV}, 2009.

\bibitem{referit}
S.~Kazemzadeh, V.~Ordonez, M.~Matten, and T.~L. Berg.
\newblock Referit game: Referring to objects in photographs of natural scenes.
\newblock In {\em EMNLP}, 2014.

\bibitem{deepaesthetics}
S.~Kong, X.~Shen, Z.~Lin, R.~Mech, and C.~Fowlkes.
\newblock {D}eep {U}nderstanding of {I}mage {A}esthetics.
\newblock In {\em ECCV}, 2016.

\bibitem{attributeshades}
A.~Kovashka and K.~Grauman.
\newblock Discovering attribute shades of meaning with the crowd.
\newblock {\em IJCV}, 114(1):56--73, 2015.

\bibitem{whittlesearch}
A.~Kovashka, D.~Parikh, and K.~Grauman.
\newblock Whittlesearch: Image search with relative attribute feedback.
\newblock In {\em CVPR}, 2012.

\bibitem{whittlesearch2}
A.~Kovashka, D.~Parikh, and K.~Grauman.
\newblock {W}hittle{S}earch: {I}nteractive {I}mage {S}earch with {R}elative
  {A}ttribute {F}eedback.
\newblock {\em IJCV}, 115(2):185--210, 2015.

\bibitem{alexnet}
A.~Krizhevsky, I.~Sutskever, and G.~Hinton.
\newblock Imagenet classification with deep convolutional neural networks.
\newblock In {\em NIPS}, pages 1097--1105, 2012.

\bibitem{babytalk}
G.~Kulkarni, V.~Premraj, S.~Dhar, S.~Li, Y.~Choi, A.~C. Berg, and T.~L. Berg.
\newblock {B}aby {T}alk: {U}nderstanding and {G}enerating {I}mage
  {D}escriptions.
\newblock In {\em CVPR}, 2011.

\bibitem{simile}
N.~Kumar, A.~C. Berg, P.~N. Belhumeur, and S.~K. Nayar.
\newblock Attribute and simile classifiers for face verification.
\newblock In {\em ICCV}, 2009.

\bibitem{betweenclassattribute}
C.~Lampert, H.~Nickisch, and S.~Harmeling.
\newblock {L}earning {T}o {D}etect {U}nseen {O}bject {C}lasses by
  {B}etween-{C}lass {A}ttribute {T}ransfer.
\newblock In {\em CVPR}, 2009.

\bibitem{deepfashion}
Z.~Liu, P.~Luo, S.~Qiu, X.~Wang, and X.~Tang.
\newblock {D}eep{F}ashion: {P}owering {R}obust {C}lothes {R}ecognition and
  {R}etrieval with {R}ich {A}nnotations.
\newblock In {\em CVPR}, 2016.

\bibitem{deeplearningface}
Z.~Liu, P.~Luo, X.~Wang, and X.~Tang.
\newblock {D}eep {L}earning {F}ace {A}ttributes in the {W}ild.
\newblock In {\em ICCV}, 2015.

\bibitem{sift}
D.~Lowe.
\newblock Distinctive image features from scale-invariant keypoints.
\newblock {\em IJCV}, 60(2):91--110, 2004.

\bibitem{partattribute}
S.~Maji and G.~Shakhnarovich.
\newblock Part and attribute discovery from relative annotations.
\newblock {\em IJCV}, 108(1):82--96, 2014.

\bibitem{unambiguousdescription}
J.~Mao, J.~Huang, A.~Toshev, O.~Camburu, A.~Yuille, and K.~Murphy.
\newblock {G}eneration and {C}omprehension of {U}nambiguous {O}bject
  {D}escriptions.
\newblock In {\em CVPR}, 2016.

\bibitem{referring}
M.~Mitchell, K.~Deemter, and E.~Reiter.
\newblock {G}enerating {E}xpressions that {R}efer to {V}isible {O}bjects.
\newblock In {\em North American Chapter of the Asssociation for Computational
  Linguistics (NAACL)}, 2013.

\bibitem{gist}
A.~Oliva and A.~Torralba.
\newblock Modeling the shape of the scene: A holistic representation of the
  spatial envelope.
\newblock {\em IJCV}, 42(3):145--175, 2001.

\bibitem{deepsaliency}
J.~Pan, E.~Sayrol, X.~Giro-i Nieto, K.~McGuinness, and N.~E. O'Connor.
\newblock Shallow and deep convolutional networks for saliency prediction.
\newblock In {\em CVPR}, 2016.

\bibitem{relativeattributes}
D.~Parikh and K.~Grauman.
\newblock Relative attributes.
\newblock In {\em ICCV}, pages 502--510, 2011.

\bibitem{sunattribute}
G.~Patterson and J.~Hays.
\newblock Sun attribute database: Discovering, annotating, and recognizing
  scene attributes.
\newblock {\em IJCV}, 108(1):59--81, 2014.

\bibitem{plattscaling}
J.~C. Platt.
\newblock Probabilistic outputs for support vector machines and comparisons to
  regularized likelihood methods.
\newblock In {\em Advances in Large Margin Classifiers}, pages 61--74. MIT
  Press, 1999.

\bibitem{spokenattributes}
A.~Sadovnik, A.~Gallagher, D.~Parikh, and T.~Chen.
\newblock Spoken attributes: Mixing binary and relative attributes to say the
  right thing.
\newblock In {\em ICCV}, 2013.

\bibitem{relativeparts}
R.~Sandeep, Y.~Verma, and C.~Jawahar.
\newblock Relative parts: Distinctive parts for learning relative attributes.
\newblock In {\em CVPR}, 2014.

\bibitem{imagerankingmulti}
B.~Siddiquie, R.~S. Feris, and L.~S. Davis.
\newblock {I}mage {R}anking and {R}etrieval based on {M}ulti-{A}ttribute
  {Q}ueries.
\newblock In {\em CVPR}, 2011.

\bibitem{deeprelative}
K.~Singh and Y.~J. Lee.
\newblock {E}nd-to-{E}nd {L}ocalization and {R}anking for {R}elative
  {A}ttributes.
\newblock In {\em ECCV}, 2016.

\bibitem{deeprelative3}
Y.~Souri, E.~Noury, and E.~Adeli.
\newblock {D}eep {R}elative {A}ttributes.
\newblock In {\em ACCV}, 2016.

\bibitem{objectimportance}
M.~Spain and P.~Perona.
\newblock {M}easuring and {P}redicting {O}bject {I}mportance.
\newblock {\em IJCV}, 91(1):59--76, 2011.

\bibitem{dominance}
N.~Turakhia and D.~Parikh.
\newblock Attribute dominance: What pops out?
\newblock In {\em ICCV}, 2013.

\bibitem{attributetoimage}
X.~Yan, J.~Yang, K.~Sohn, and H.~Lee.
\newblock {A}ttribute2{I}mage: {C}onditional {I}mage {G}eneration from {V}isual
  {A}ttributes.
\newblock In {\em ECCV}, 2016.

\bibitem{deeprelative2}
X.~Yang, T.~Zhang, C.~Xu, S.~Yan, M.~S. Hossain, and A.~Ghoneim.
\newblock Deep relative attributes.
\newblock {\em IEEE Transactions on Multimedia}, 18(9):1832--1842, 2016.

\bibitem{finegrained}
A.~Yu and K.~Grauman.
\newblock {F}ine-{G}rained {V}isual {C}omparisons with {L}ocal {L}earning.
\newblock In {\em CVPR}, 2014.

\bibitem{jnd}
A.~Yu and K.~Grauman.
\newblock Just noticeable differences in visual attributes.
\newblock In {\em ICCV}, 2015.

\bibitem{semanticjitter}
A.~Yu and K.~Grauman.
\newblock Semantic jitter: Dense supervision for visual comparisons via
  synthetic images.
\newblock In {\em ICCV}, 2017.

\bibitem{fashionsearch}
B.~Zhao, J.~Feng, X.~Wu, and S.~Yan.
\newblock {M}emory-{A}ugmented {A}ttribute {M}anipulation {N}etworks for
  {I}nteractive {F}ashion {S}earch.
\newblock In {\em CVPR}, 2017.

\end{thebibliography}

\end{document}